\documentclass[11pt]{article}

\usepackage[final]{acl-style-files-master/acl}

\usepackage{times}
\usepackage{latexsym}

\usepackage[T1]{fontenc}

\usepackage[utf8]{inputenc}

\usepackage{microtype}

\usepackage{inconsolata}

\usepackage{graphicx}
\usepackage{booktabs}

\title{Best of Both Worlds: Fine-Tuned Small Models Complement LLMs for Cost and Performance Benefits}

\author{Shivam Mathur\thanks{Equal contribution, names listed in alphabetical order.}\thanks{Corresponding author: \texttt{shivam.mathur@walmart.com}} \And
  Sri Harish Popuri\footnotemark[1] \And
  Priya Krishnan Sundararajan\footnotemark[1]\thanks{Work done during time at Walmart Inc.} \AND
  Kevin Horecka \And
  Michael Mugo \AND
  {\normalfont Walmart Inc.}
  }

\begin{document}
\maketitle
\begin{abstract}
BERT-based models require thousands of labeled samples to train. 
Larger models like GPT--X require only 1--5 examples per class. 
However, performance, cost and time tradeoffs of such models remain under-explored. 
In this work, we classify survey text data into 16 classes. 
We focus on experimental approaches and results on performance, cost, and time taken for this problem. 
We compare these parameters of a trained RoBERTa-based model, and an (off-the-shelf) GPT--3.5 model for text classification. 
Baseline experiments show that the time taken by GPT--3.5 for generating predictions on 10,000 records is 6 hours, whereas a RoBERTa-based model takes only 12 mins on a T4 GPU. 
To identify the best prompting approach for GPT--3.5, we explore batch prompting. 
We experiment with (1) batch-size, and (2) few-shot sampling. 
We find the best batch size is 20 with human few-shot curation. 
Experiments show the F1 score of a trained RoBERTa-based model (96\%) is much higher than a GPT--3.5 model (83\%). 
In case of GPT--3.5, batch prompting is beneficial in lowering inference costs by ~70\%, without hindering task performance. 
We find the best approach is an ensemble of a RoBERTa-based and GPT--3.5 model (F1: 97.5\%) incurring only 4.4\% of inference cost compared to a standalone GPT--3.5. 
\end{abstract}

\section{Introduction}
\label{s:introduction}

Natural language processing has seen impressive advancements in the past few years, especially in text classification. 
Large language models have achieved performance comparable to supervised baselines such as 
question answering \citep{trivedi-etal-2023-interleaving}, 
named entity recognition \citep{wang-etal-2025-gpt}, 
relation extraction \citep{wan-etal-2023-gpt} and
human annotator performance \citep{Gilardi_2023}. 
\\
Despite the remarkable success, previous studies \citep{sun-etal-2023-text, mathur-etal-2023-interpreting}
show that LLMs may still significantly underperform fine-tuned models in the task of text classification. 
In this paper, we do a performance, cost, and time tradeoff of GPT--3.5 and a supervised RoBERTa-based model on text classification task. 
We experiment with different batch sizes, few shot sizes and sample selection strategies to identify the best prompting approach in terms of performance, cost, and time suitable for the dataset. 
The dataset is a survey with an open text response. 
We apply NLP models to classify these response texts into pre-defined topics/classes for better insights and speedy actions.
\\
Our final approach involves using a combination of fine-tuned RoBERTa-based model and GPT--3.5, where GPT--3.5 is used to classify the response texts that have a confidence less than 0.5. 
This approach boosts F1 score to 97.5\% incurring only 4.4\% of inference cost compared to when an off-the-shelf LLM (GPT--3.5) alone was used.  

\section{Problem Background}
\label{s:background}

Open text response classification is challenging. 
It involves reasoning over long text and overlapping diverse topics. 
For experiments in this work, we consider a subset of survey text responses (around 10,000 samples) for prediction and 30,000 of labeled texts for generating few shot samples or training. 
We pick 16 topics with clear definitions that are mutually exclusive.
\section{Experimental Setup}
\label{s:expt_setup}
We compare the F1 score, time, and cost of a fine-tuned RoBERTa-based model and an off-the-shelf GPT--3.5 model under ideal batch size and few shot sample configuration. 
To identify the optimal configuration of batch size and few shot samples, we have the below described experiment setup. 
All experiments are run using the Element platform's Airflow jobs, and a Tesla T4 GPU to do training and inferences.

\subsection{Baseline Experiments}
\begin{table}[t]
\centering
\begin{tabular}{lrrrr}
\toprule
Model & F1 & \multicolumn{2}{c}{Time} & Total Cost \\
\cmidrule(rr){3-4}
 & & train & inf. & \\
\midrule
RoBERTa  & 96  & 106  & 12  & 11K  \\
\multicolumn{5}{l}{GPT--3.5} \\
~~~Zero-shot  & 79  & N/A  & 340  & 16  \\
~~~Few-shot  & 79  & N/A  & 371  & 44  \\
\bottomrule
\end{tabular}
\caption{Baseline models with F1-score, training and prediction (inference) time (minutes), and total cost (\$).
Training is done on 21,000 samples and prediction is done on 10,000 samples.
Cost is an estimate of human annotation effort involved in the annotation task. 
We use Manthano.ai (a free tool) for human annotation cost estimations (in 2023). Costs for GPT--3.5 are only inference costs (no human annotation needed).
}
\label{t:baselines}
\end{table}
\label{s:baselines}
Our baseline experiments consist of three models (Table~\ref{t:baselines}). 
The first model is a fine-tuned RoBERTa-based model. 
It is trained on 21,000 samples and validated on 9,000 samples. 
The second is a zero-shot GPT--3.5 model and the final model is a 3-few-shot GPT--3.5. 
The RoBERTa-based model outperforms both GPT--3.5 approaches.

\subsection{Batch Prompting}

\label{s:batch_prompting}
In the preliminary stage of our experiment, we seek to determine the optimal batch size for our input data. 
To maintain consistency, we kept the size of few-shot examples constant (3 samples) while varying the batch sizes from 10 to 50 in increments of 10. 
This approach enables conducting an exploration of different batch sizes and its impact on the overall performance, time and cost.
This also served as a step in deciding the optimal batch size for subsequent few shot experiments. 

\subsection{Few-Shot Techniques}
\label{s:few_shot_techniques}
For few-shot experiments, we use samples that are only one or two sentences long as few shot examples. 
Longer response texts can have two or more topics associated with them introducing complexity for the model. 
Also, having shorter few-shot examples keeps runtimes and costs within budget. 
The two strategies for few-shot selection are (1) random selection and (2) human-curated selection. 
Batch size is set constant to 20, which found to be optimal from batch prompting experiments. 
We vary the number of few shot examples from one to nine, in increments of two.

\section{Experimental Results}
\label{s:experimental_results}

\subsection{Batch Prompting}
\label{s:batch_prompting_results}
\begin{figure}
\graphicspath{{./figures/images/}}
\includegraphics[width=\columnwidth]{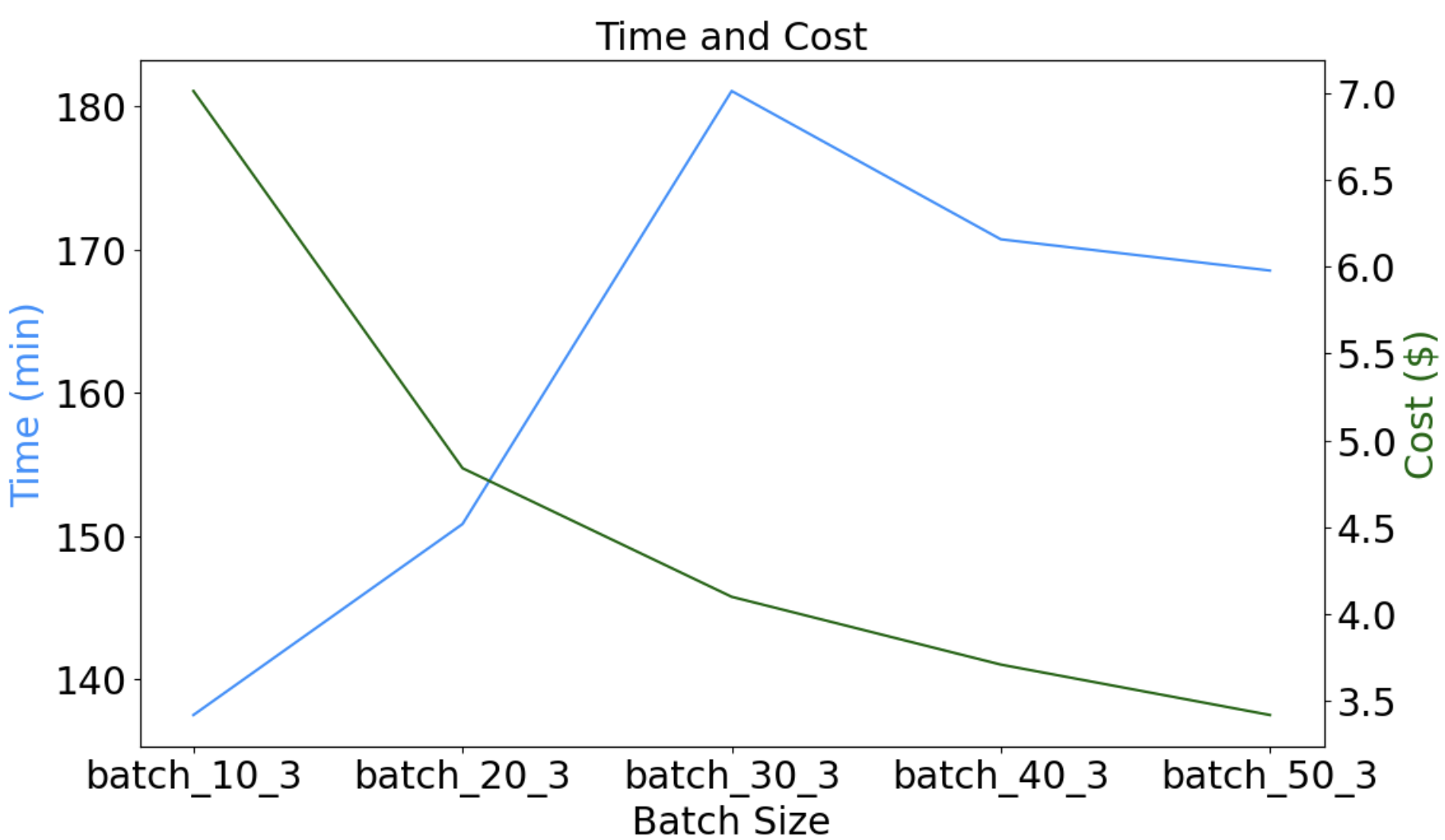}
\caption{Variations in total time (minutes) and cost (\$) observed across experiments with batch sizes ranging from 10 to 50 (fewshot kept constant). 
X-axis labels are annotated as \emph{batch\_[batch size]\_[few-shot count]}.
Larger batch sizes are more cost-effective but take longer time (in most cases).
}
\label{f:batches_time_cost}
\end{figure}
\begin{figure}
\graphicspath{{./figures/images}}
\includegraphics[width=\columnwidth]{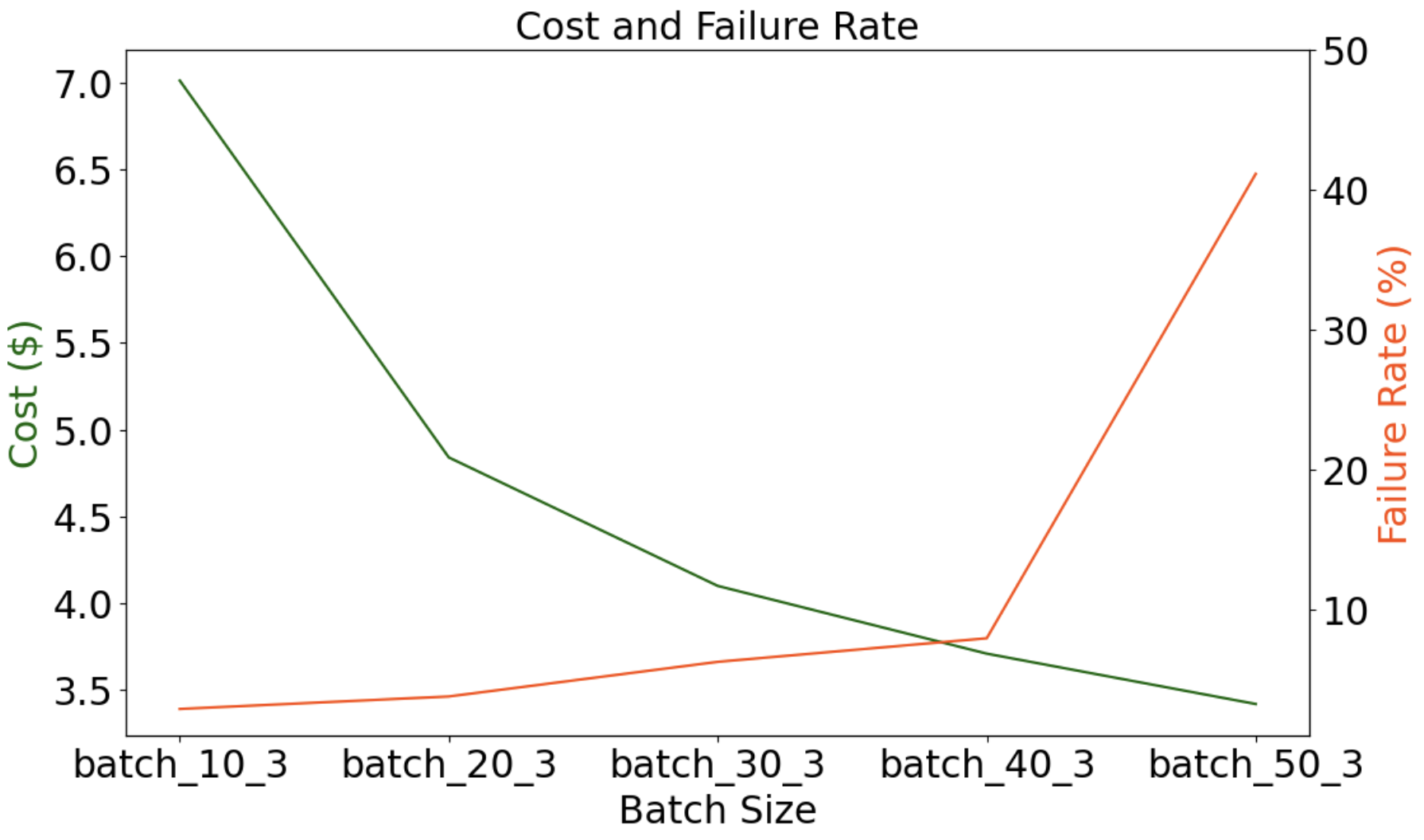}
\caption{Variations in total cost (\$) and failure (\%) observed across experiments with batch sizes ranging from 10 to 50 (fewshot kept constant). 
X-axis labels are annotated as \emph{batch\_[batch size]\_[few-shot count]}.
Larger batch sizes are more cost-effective (similar to Figure~\ref{f:batches_time_cost}) but lead to higher failure rate.
}
\label{f:batches_cost_failure}
\end{figure}
\begin{table}[t]
\centering
\begin{tabular}{lrrrrr}
\toprule
Batch Size & 10 & 20 & 30 & 40 & 50 \\
\midrule
F1 (\%)       & 82  & 82  & 82  & 82  & 83   \\
Time (min)    & 138 & 151 & 181 & 171 & 168  \\
Cost (\$)     & 7   & 4.8 & 4.1 & 3.7 & 3.4  \\
Failure (\%)  & 2.9 & 3.8 & 6.3 & 8   & 41.5 \\
\bottomrule
\end{tabular}
\caption{Batch size vs. F1, time, cost and failure \%. 
Optimal batch size is 20 which keeps time, cost and failures rate low while maintaining the F1 score.
It is also 70\% more cost-effective compared to the GPT--3.5 zero-shot baseline (\$16 in Table~\ref{t:baselines}).}
\label{t:batch_f1}
\end{table}

An optimal batch size is identified by examining the outcomes across three dimensions: 
Time, cost, and failure rate (percentage of instances where the model failed to generate any prediction).
We vary batch sizes (10, 20, 30, 40, 50) to attempt to optimize cost relative to the risk of batches failing due to being too large for the LLM context window (Figure~\ref{f:batches_time_cost}). 
There is no significant change in performance across batch size (Table~\ref{t:batch_f1}), but even in the presence of a sharp increase (2.9 \% to 41.5 \%) in batch failures (Figure~\ref{f:batches_cost_failure}), 
using larger batch sizes still results in the best cost (\$3.4 for batch size 50). 

\subsection{Few-Shot Selection Techniques}
\label{s:few_shot_selection_results}
\begin{table}[t]
\centering
\begin{tabular}{lrrrrr}
\toprule
Few-shot \# & 1 & 3 & 5 & 7 & 9 \\
\midrule
\multicolumn{6}{l}{Random-selected} \\
~~~F1 (\%)       & 81  & 82  & 83  & 83  & 83   \\
~~~Time (min)    & 159 & 172 & 190 & 205 & 201  \\
~~~Cost (\$)     & 3.7 & 4.8 & 6.1 & 7.1 & 8.3  \\
~~~Failure \%    & 4.5 & 5.0 & 4.1 & 4.0 & 5.3 \\
\midrule
\multicolumn{6}{l}{Human-curated} \\
~~~F1(\%)        & 82  & 82  & 83  & 83  & 83   \\
~~~Time (min)    & 226 & 304 & 299 & 238 & 207  \\
~~~Cost (\$)     & 3.4 & 4.0 & 4.5 & 5.0 & 5.6  \\
~~~Failure \%    & 6.9 & 4.2 & 4.9 & 5.7 & 4.5 \\
\bottomrule
\end{tabular}
\caption{Few-shot size vs F1, time, cost \& failure \%.
Human-curated few-shot examples are more cost-effective (due to shorter token counts (Table~\ref{t:cost_token_length})).}
\label{t:few_shot_f1}
\end{table}

\begin{table}[t]
\centering
\small
\begin{tabular}{lrrrr}
\toprule
 & \multicolumn{3}{c}{Mean token count} & \\
\cmidrule(lr){2-4}
Selection & Prompt & Generation & Reasoning & \begin{tabular}[r]{@{}r@{}}Cost \end{tabular} \\
\midrule
\multicolumn{5}{l}{Few-shot 1} \\
~~~Random & 1059.4 & 1043.2 & 18.4 & 3.7 \\
~~~Human  & 910.4  & 1033.7 & 18.2 & 3.5 \\
\midrule
\multicolumn{5}{l}{Few-shot 3} \\
~~~Random & 1794.4 & 1053.9 & 19.0 & 4.9 \\
~~~Human  & 1257.4 & 1046.5 & 18.7 & 4.0 \\
\midrule
\multicolumn{5}{l}{Few-shot 5} \\
~~~Random & 2620.4 & 1059.2 & 19.3 & 6.1 \\
~~~Human  & 1605.4 & 1049.3 & 18.8 & 4.6 \\
\midrule
\multicolumn{5}{l}{Few-shot 7} \\
~~~Random & 3259.4 & 1071.0 & 19.9 & 7.1 \\
~~~Human  & 1930.4 & 1057.0 & 19.2 & 5.1 \\
\midrule
\multicolumn{5}{l}{Few-shot 9} \\
~~~Random & 4047.4 & 1074.0 & 20.1 & 8.3 \\
~~~Human  & 2279.4 & 1062.8 & 19.5 & 5.6 \\
\bottomrule
\end{tabular}
\caption{Cost (\$) and token count comparison of random-selected and human-curated few-shot experiments in various stages (prompt, generation and reasoning). 
Human-curated experiments are more cost efficient.
This table complements Table~\ref{t:few_shot_f1}.}
\label{t:cost_token_length}
\end{table}

Table~\ref{t:few_shot_f1} shows the results of random and human-curated few shot sample selections. 
We test various few-shot sample sizes (1, 3, 5, 7, 9) using both a random-selected and human-curated method. 
Costs increase with sample-size in the random condition faster than in the human-condition.
This is largely driven by the fact that human-curated examples tend to have shorter token counts 
(more succinct) 
(Table~\ref{t:cost_token_length}: 4047 vs. 2279 in the 9-example condition). 
\section{Related Work}
\label{s:related_work}

\subsection{Batch Prompting Techniques}
\label{s:rel_batch_prompting_techniques}
Batch prompting, an efficient way to prompt LLMs. 
It performs inference on samples in a batched fashion.
Multiple samples can be handled in one API call so that the costs of tokens and time can be significantly reduced \citep{cheng-etal-2023-batch}.
Also, experiments with different permutations of input samples for every batch might affect performance and time metrics. 
This involves a voting method to finally determine the prediction, to minimize the variance in final outcomes and controlling indefinite rounds of voting \citep{ICLR2024_5d8c01de}.  

\subsection{Few-Shot Techniques}
\label{s:rel_few_shot_techniques}
In-context learning involves providing a few demonstration examples in the prompt template. 
Performance of GPT--3.5 can show quite the dependence on a few shot examples \citep{NEURIPS2020_1457c0d6}. 
There has been past work in determining ways to select such examples for attaining better performance.
Skill-based k-nearest neighbor approaches have been used to select few shot examples \citep{an-etal-2023-skill}. 
Trained retriever models \citep{rubin-etal-2022-learning}
are also used to extract the best suited examples for tasks.  
\section{Conclusion}
We carry out an exploration of performance, cost and time tradeoff of a fine-tuned RoBERTa-based model and a GPT--3.5 model for a text classification task. 
We explore batch prompting, a technique where multiple responses can be processed in a single API call thus saving cost and time. 
Upon experimenting with different batch sizes, we find that a batch size of 20 is optimal in terms of performance, cost and time for our data. 
Human-curation of few-shot examples emerges as a cost-effective strategy, with an optimal size of around three examples to achieve a favorable F1-score without incurring unnecessary costs. 
In Appendix~\ref{s:appendix}, we examine errors, exploring error types such as compound, domain-based, and granularity-based errors, highlighting the need for labels and few-shot examples to be unambiguous and mutually exclusive. 
Our final approach of using an ensemble model by first using RoBERTa-based model and then using GPT--3.5 on the records where former is less confident achieves superior performance of 97.5\% incurring only 4.4\% of inference cost compared to when a pre-trained LLM (GPT--3.5) alone was used. 
For generating training data, we propose leveraging LLMs along with human labeled data. 
As future work, we would like to explore batching similar records and other batch selection techniques.
Also, exploring current state-of-art LLMs and other open-source LLMs to assess the trade-offs would be a promising direction. 

\section*{Acknowledgments}
We would like to thank Deeksha Yennam and Punith Krishna Pothula for their thoughtful reviews, suggestions, and immense support over time. 
We would also like to acknowledge the Walmart review board for their support. 

\bibliography{custom}

@inproceedings{trivedi-etal-2023-interleaving,
    title = "Interleaving Retrieval with Chain-of-Thought Reasoning for Knowledge-Intensive Multi-Step Questions",
    author = "Trivedi, Harsh  and
      Balasubramanian, Niranjan  and
      Khot, Tushar  and
      Sabharwal, Ashish",
    editor = "Rogers, Anna  and
      Boyd-Graber, Jordan  and
      Okazaki, Naoaki",
    booktitle = "Proceedings of the 61st Annual Meeting of the Association for Computational Linguistics (Volume 1: Long Papers)",
    month = jul,
    year = "2023",
    address = "Toronto, Canada",
    publisher = "Association for Computational Linguistics",
    url = "https://aclanthology.org/2023.acl-long.557/",
    doi = "10.18653/v1/2023.acl-long.557",
    pages = "10014--10037"
}

@inproceedings{wang-etal-2025-gpt,
    title = "{GPT}-{NER}: Named Entity Recognition via Large Language Models",
    author = "Wang, Shuhe  and
      Sun, Xiaofei  and
      Li, Xiaoya  and
      Ouyang, Rongbin  and
      Wu, Fei  and
      Zhang, Tianwei  and
      Li, Jiwei  and
      Wang, Guoyin  and
      Guo, Chen",
    editor = "Chiruzzo, Luis  and
      Ritter, Alan  and
      Wang, Lu",
    booktitle = "Findings of the Association for Computational Linguistics: NAACL 2025",
    month = apr,
    year = "2025",
    address = "Albuquerque, New Mexico",
    publisher = "Association for Computational Linguistics",
    url = "https://aclanthology.org/2025.findings-naacl.239/",
    doi = "10.18653/v1/2025.findings-naacl.239",
    pages = "4257--4275",
    ISBN = "979-8-89176-195-7"
}

@inproceedings{wan-etal-2023-gpt,
    title = "{GPT}-{RE}: In-context Learning for Relation Extraction using Large Language Models",
    author = "Wan, Zhen  and
      Cheng, Fei  and
      Mao, Zhuoyuan  and
      Liu, Qianying  and
      Song, Haiyue  and
      Li, Jiwei  and
      Kurohashi, Sadao",
    editor = "Bouamor, Houda  and
      Pino, Juan  and
      Bali, Kalika",
    booktitle = "Proceedings of the 2023 Conference on Empirical Methods in Natural Language Processing",
    month = dec,
    year = "2023",
    address = "Singapore",
    publisher = "Association for Computational Linguistics",
    url = "https://aclanthology.org/2023.emnlp-main.214/",
    doi = "10.18653/v1/2023.emnlp-main.214",
    pages = "3534--3547"
}

@article{Gilardi_2023,
   title={ChatGPT outperforms crowd workers for text-annotation tasks},
   volume={120},
   ISSN={1091-6490},
   url={http://dx.doi.org/10.1073/pnas.2305016120},
   DOI={10.1073/pnas.2305016120},
   number={30},
   journal={Proceedings of the National Academy of Sciences},
   publisher={National Academy of Sciences},
   author={Gilardi, Fabrizio and Alizadeh, Meysam and Kubli, Maël},
   year={2023},
   month=jul 
}

@inproceedings{sun-etal-2023-text,
    title = "Text Classification via Large Language Models",
    author = "Sun, Xiaofei  and
      Li, Xiaoya  and
      Li, Jiwei  and
      Wu, Fei  and
      Guo, Shangwei  and
      Zhang, Tianwei  and
      Wang, Guoyin",
    editor = "Bouamor, Houda  and
      Pino, Juan  and
      Bali, Kalika",
    booktitle = "Findings of the Association for Computational Linguistics: EMNLP 2023",
    month = dec,
    year = "2023",
    address = "Singapore",
    publisher = "Association for Computational Linguistics",
    url = "https://aclanthology.org/2023.findings-emnlp.603/",
    doi = "10.18653/v1/2023.findings-emnlp.603",
    pages = "8990--9005"
}

@inproceedings{cheng-etal-2023-batch,
    title = "Batch Prompting: Efficient Inference with Large Language Model {API}s",
    author = "Cheng, Zhoujun  and
      Kasai, Jungo  and
      Yu, Tao",
    editor = "Wang, Mingxuan  and
      Zitouni, Imed",
    booktitle = "Proceedings of the 2023 Conference on Empirical Methods in Natural Language Processing: Industry Track",
    month = dec,
    year = "2023",
    address = "Singapore",
    publisher = "Association for Computational Linguistics",
    url = "https://aclanthology.org/2023.emnlp-industry.74/",
    doi = "10.18653/v1/2023.emnlp-industry.74",
    pages = "792--810"
}

@inproceedings{ICLR2024_5d8c01de,
 author = {Lin, Jianzhe and Diesendruck, Maurice and Du, Liang and Abraham, Robin},
 booktitle = {International Conference on Learning Representations},
 editor = {B. Kim and Y. Yue and S. Chaudhuri and K. Fragkiadaki and M. Khan and Y. Sun},
 pages = {21590--21612},
 title = {BatchPrompt: Accomplish more with less},
 url = {https://proceedings.iclr.cc/paper_files/paper/2024/file/5d8c01de2dc698c54201c1c7d0b86974-Paper-Conference.pdf},
 volume = {2024},
 year = {2024}
}

@inproceedings{NEURIPS2020_1457c0d6,
 author = {Brown, Tom and Mann, Benjamin and Ryder, Nick and Subbiah, Melanie and Kaplan, Jared D and Dhariwal, Prafulla and Neelakantan, Arvind and Shyam, Pranav and Sastry, Girish and Askell, Amanda and Agarwal, Sandhini and Herbert-Voss, Ariel and Krueger, Gretchen and Henighan, Tom and Child, Rewon and Ramesh, Aditya and Ziegler, Daniel and Wu, Jeffrey and Winter, Clemens and Hesse, Chris and Chen, Mark and Sigler, Eric and Litwin, Mateusz and Gray, Scott and Chess, Benjamin and Clark, Jack and Berner, Christopher and McCandlish, Sam and Radford, Alec and Sutskever, Ilya and Amodei, Dario},
 booktitle = {Advances in Neural Information Processing Systems},
 editor = {H. Larochelle and M. Ranzato and R. Hadsell and M.F. Balcan and H. Lin},
 pages = {1877--1901},
 publisher = {Curran Associates, Inc.},
 title = {Language Models are Few-Shot Learners},
 url = {https://proceedings.neurips.cc/paper_files/paper/2020/file/1457c0d6bfcb4967418bfb8ac142f64a-Paper.pdf},
 volume = {33},
 year = {2020}
}

@inproceedings{an-etal-2023-skill,
    title = "Skill-Based Few-Shot Selection for In-Context Learning",
    author = "An, Shengnan  and
      Zhou, Bo  and
      Lin, Zeqi  and
      Fu, Qiang  and
      Chen, Bei  and
      Zheng, Nanning  and
      Chen, Weizhu  and
      Lou, Jian-Guang",
    editor = "Bouamor, Houda  and
      Pino, Juan  and
      Bali, Kalika",
    booktitle = "Proceedings of the 2023 Conference on Empirical Methods in Natural Language Processing",
    month = dec,
    year = "2023",
    address = "Singapore",
    publisher = "Association for Computational Linguistics",
    url = "https://aclanthology.org/2023.emnlp-main.831/",
    doi = "10.18653/v1/2023.emnlp-main.831",
    pages = "13472--13492"
}

@inproceedings{rubin-etal-2022-learning,
    title = "Learning To Retrieve Prompts for In-Context Learning",
    author = "Rubin, Ohad  and
      Herzig, Jonathan  and
      Berant, Jonathan",
    editor = "Carpuat, Marine  and
      de Marneffe, Marie-Catherine  and
      Meza Ruiz, Ivan Vladimir",
    booktitle = "Proceedings of the 2022 Conference of the North American Chapter of the Association for Computational Linguistics: Human Language Technologies",
    month = jul,
    year = "2022",
    address = "Seattle, United States",
    publisher = "Association for Computational Linguistics",
    url = "https://aclanthology.org/2022.naacl-main.191/",
    doi = "10.18653/v1/2022.naacl-main.191",
    pages = "2655--2671"
}

@inproceedings{mathur-etal-2023-interpreting,
    title = "Interpreting Answers to Yes-No Questions in User-Generated Content",
    author = "Mathur, Shivam  and
      Park, Keun  and
      Chinnappa, Dhivya  and
      Kotamraju, Saketh  and
      Blanco, Eduardo",
    editor = "Bouamor, Houda  and
      Pino, Juan  and
      Bali, Kalika",
    booktitle = "Findings of the Association for Computational Linguistics: EMNLP 2023",
    month = dec,
    year = "2023",
    address = "Singapore",
    publisher = "Association for Computational Linguistics",
    url = "https://aclanthology.org/2023.findings-emnlp.942/",
    doi = "10.18653/v1/2023.findings-emnlp.942",
    pages = "14137--14161"
}

\appendix
\section{Appendix 1: Error Analysis}
\label{s:appendix}
\begin{table*}[t]
\centering
\small
\begin{tabular}{p{0.45\columnwidth}p{0.6\columnwidth}p{0.30\columnwidth}p{0.25\columnwidth}}
\toprule
Error (\%) & Example & GPT--3.5 & RoBERTa/Gold \\
\midrule
\textit{Compound} (23\%) & Train on the equipment & Equipment & Development \\
\textit{Domain / Short Text} (35\%) & Snack & No Complaint & Benefits \\
\textit{Granularity} (42\%) & Info on hours in the assigned dept & Communicate & Shifts \\
\bottomrule
\end{tabular}
\caption{Analysis of GPT--3.5 errors compared to correct RoBERTa-based and Gold on a random sample of 100 instances. 
Examples are paraphrased and hence synthetic. 
This table complements Appendix~\ref{s:appendix}.}
\label{t:error_analysis}
\end{table*}

We analyze instances where the RoBERTa-based model's predicted class is correct, but GPT--3.5 predicted class is incorrect. 
This analysis is done on a random sample of 100 error instances.
We human-evaluate for error classes. 
This approach seeks inspiration from \citep{mathur-etal-2023-interpreting}. 
We find three main error groups shown in Table~\ref{t:error_analysis}. 
Text examples mentioned are paraphrased and hence synthetic. 

\textbf{Compound}:
The response text may contain tokens of a certain class (“equipment”) but belongs to some other class. 
For example, “Train on the equipment” ought to fall into Development as opposed to Equipment (since it talks about training).

\textbf{Domain or Short Text}:
An off-the-shelf GPT--3.5 despite having provided few-shot examples lacks the deep domain-level understanding of a trained RoBERTa-based model. 
Especially in cases where response texts can be shorter (giving lesser context). 
In the example, “snack” refers to the provision of snacks, which is aligned to benefits, as opposed to no complaint.

\textbf{Granularity}:
At times GPT--3.5 assigns a class, but its reasons towards a rather more generic class. 
In the example “Info on hours” does imply some level of communication but more specifically is a shift related topic.  
\end{document}